\documentclass{article}
\usepackage{spconf,amsmath,amssymb,graphicx,hyperref,enumitem}
\hypersetup{hidelinks}
\title{TCNERV: DUAL-DOMAIN TEMPORAL CONTEXT MODELING FOR IMPLICIT NEURAL VIDEO COMPRESSION}

\name{\normalsize Xuezhi Xiang$^{1}$$^{,}$$^{2}$$^{,}$$^{*}$ \quad Yixin Zhao$^{1}$ \quad
	Heqi Xiang$^{3}$ \quad Jiayao Liu$^{1}$ \quad
	Shanjun Zhang$^{4}$
	\thanks{This work was supported in part by the National Natural Science Foundation of China under Grants 62671193 and 62271160, in part by the Heilongjiang Provincial Key R\&D Program under Grant 2026ZX01A14, in part by the Natural Science Foundation of Heilongjiang Province of China under Grant XQ2026F018, and in part by the Fundamental Research Funds for the Central Universities of China under Grant 3072026LJ0802.}}
\address{\normalsize $^{1}$Information and Communication Engineering, Harbin Engineering University, Harbin, 150001, China\\
	\normalsize $^{2}$Key Laboratory of Advanced Marine Communication and Information Technology, Harbin, 150001, China\\
	\normalsize $^{3}$Department of Computer Science, University of Toronto, Toronto, ON M5S 2E4, Canada\\
	\normalsize $^{4}$The Department of Computer Science, Kanagawa University, Kanagawa, 221-8686, Japan\\	
	\normalsize \href{mailto:xiangxuezhi@hrbeu.edu.cn}{xiangxuezhi@hrbeu.edu.cn};
	\href{mailto:yixinzhao@hrbeu.edu.cn}{yixinzhao@hrbeu.edu.cn}\\
	\normalsize \href{mailto:claire.xiang@mail.utoronto.ca}{claire.xiang@mail.utoronto.ca};
	\href{mailto:liujiayao@hrbeu.edu.cn}{liujiayao@hrbeu.edu.cn};
	\href{mailto:chiyoz01@kanagawa-u.ac.jp}{chiyoz01@kanagawa-u.ac.jp}}
	
\begin{document}
\maketitle

\begin{abstract}
Video compression aims to minimize reconstruction distortion under a constrained bit rate. Existing video implicit neural representations (INRs) often decode frames independently, leaving intermediate features unconditioned on previous reconstructions and content embeddings without explicit temporal prediction. We propose TCNeRV, which exploits reconstructed context in both feature and embedding domains. Its multi-scale temporal-context fusion (MTCF) module injects gated historical features at multiple decoder scales, while temporal embedding-residual coding (TERC) predicts each content embedding and codes only its residual. With approximately 3M parameters, TCNeRV achieves an average PSNR of 36.08 dB on the UVG dataset, outperforming HNeRV-Boost by 2.20 dB. It reduces BD-rate by 22.06\%, 66.73\%, and 29.85\% relative to HM, DCVC, and HiNeRV, respectively, demonstrating competitive rate-distortion performance with limited model capacity.
\end{abstract}


\section{Introduction}
\label{sec:introduction}

Video compression minimizes reconstruction distortion under a constrained bit rate. Conventional standards, including HEVC~\cite{sullivan2012hevc} and VVC~\cite{bross2021vvc}, improve coding efficiency through block partitioning, prediction, transform quantization, and entropy coding. Learning-based codecs jointly optimize major coding components. DVC~\cite{lu2019dvc} introduced end-to-end motion and residual coding, DCVC~\cite{li2021dcvc} adopted conditional coding with temporal features, and DCVC-DC~\cite{li2023diverse} employed diverse spatiotemporal contexts. However, their multiple motion, context, residual, and entropy-modeling sub-networks incur considerable complexity.

Implicit neural representations (INRs) instead parameterize signals using coordinate- or index-conditioned networks. Periodic activations~\cite{sitzmann2020siren} improve high-frequency fitting, while COIN~\cite{dupont2021coin} enables content-specific image compression. For video, NeRV~\cite{chen2021nerv} maps frame indices directly to frames, E-NeRV~\cite{li2022enerv} disentangles spatial-temporal contexts, and HNeRV~\cite{chen2023hnerv} introduces content-adaptive embeddings. HiNeRV~\cite{kwan2023hinerv} uses hierarchical multi-scale encoding, and HNeRV-Boost~\cite{zhang2024hnervboost} improves feature-to-frame alignment through conditional decoding and temporal-aware modulation. Nevertheless, these methods largely represent frames independently. Temporal INR methods include FFNeRV~\cite{lee2023ffnerv}, which propagates information using optical flow, and DNeRV~\cite{zhao2023dnerv}, which separates spatial content and frame differences. Such mechanisms require additional branches, while unselective temporal fusion may propagate mismatched features under fast motion, occlusion, or scene changes.

INR compression further requires quantization and entropy coding. Early work compressed video-specific networks~\cite{zhang2022implicit}, while entropy-constrained representations~\cite{gomes2023entropy} incorporated rate estimation. C3~\cite{kim2024c3} targets efficient decoding, NVRC~\cite{kwan2024nvrc} jointly optimizes representations, quantization, and entropy models, and GIViC~\cite{gao2025givic} exploits generative priors and long-range dependencies. However, hybrid video INRs still encode frame-wise embeddings independently.

\begin{figure*}[t]
	\centering
	\includegraphics[width=\textwidth]{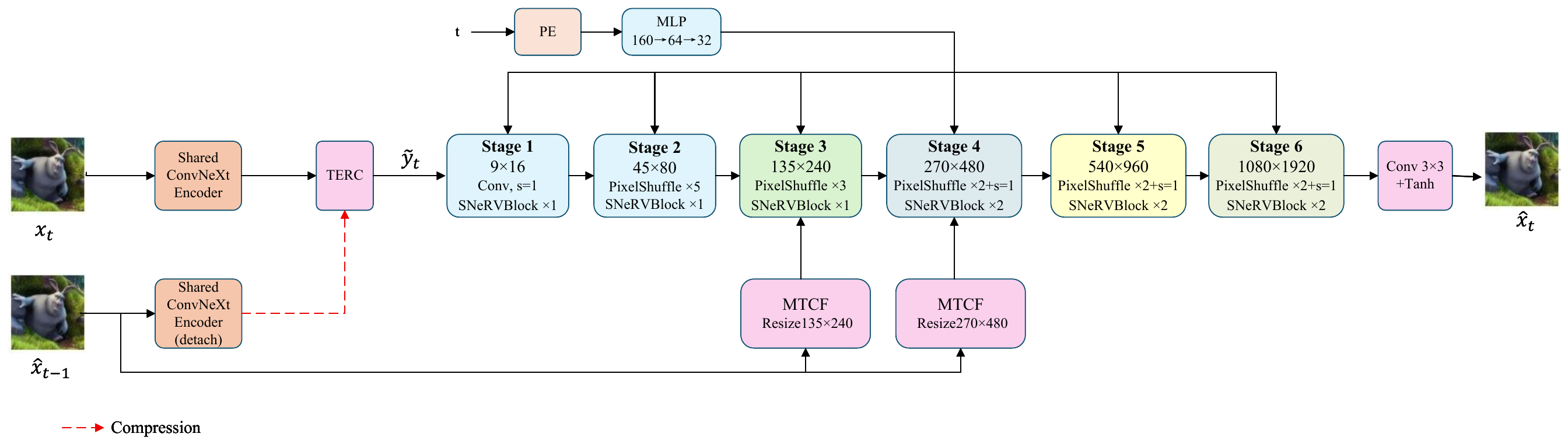}
	\caption{Overall architecture of TCNeRV. MTCF enhances current-frame structure and texture reconstruction through multi-scale feature extraction and channel-wise gating; TERC predicts the current content embedding from historical reconstruction context and establishes an explicit temporal predictive-coding relationship among framewise content embeddings.}
	\label{fig:overall}
\end{figure*}

Existing approaches therefore have two limitations: intermediate decoding features cannot reuse structures and textures from previous reconstructions, and independently coded embeddings cannot remove temporally predictable information before quantization.

Inspired by the gated context selection of BiECVC~\cite{jiang2025biecvc}, we propose multi-scale temporal-context fusion (MTCF), which aligns features from the previous reconstruction with two decoder stages and integrates them through channel-wise gated residual connections. Inspired by the autoregressive representation of NIRVANA~\cite{maiya2023nirvana}, we further propose temporal embedding-residual coding (TERC), which predicts the current embedding from reconstructed context and encodes only its residual. MTCF reduces distortion in the decoding-feature domain, while TERC exploits redundancy in the embedding domain. Together, they constitute TCNeRV.

Our contributions are summarized as follows:
\begin{enumerate}[nosep]
  \item We propose TCNeRV, which constructs causal temporal context from the previous reconstructed frame and exploits inter-frame correlation in both decoding features and content embeddings.
  \item We propose MTCF, which performs gated residual fusion at two decoder scales to improve structure and texture reconstruction without explicit motion estimation.
  \item We propose TERC, which replaces full-embedding coding with frame-level embedding-prediction-residual coding before quantization.
  \item Experiments on the UVG dataset validate TCNeRV. For PSNR-oriented compression, it reduces BD-rate by 22.06\%, 66.73\%, and 29.85\% relative to HM, DCVC, and HiNeRV, respectively.
\end{enumerate}

\section{Method}
\label{sec:method}

\subsection{Method Overview}
\label{sec:overview}

TCNeRV builds on HNeRV-Boost~\cite{zhang2024hnervboost} and represents a video using quantized content embeddings, temporal embeddings, and conditional-decoder parameters. As shown in Fig.~\ref{fig:overall}, a shared ConvNeXt encoder~\cite{liu2022convnext} extracts the current content embedding. TERC predicts this embedding from the previous reconstruction and encodes only the prediction residual, thereby exploiting temporal redundancy. The recovered content and temporal embeddings are fed into a six-stage conditional decoder, where Stage 1 expands the channels and Stages 2--6 progressively restore spatial resolution. At Stages 3 and 4, MTCF aligns the historical context with the current decoding features and performs gated fusion. The first frame uses zero context, while subsequent frames use the preceding reconstruction, forming a causal closed-loop process shared by the encoder and decoder.

\subsection{Multi-Scale Temporal-Context Fusion (MTCF)}
\label{sec:mtcf}

Adjacent frames share structures and textures, but historical cues may become unreliable under occlusion, fast motion, or scene changes. MTCF therefore extracts context at two decoder scales and integrates it through channel-wise gates.

As shown in Fig.~\ref{fig:mtcf}, MTCF operates at Stages 3 and 4 with resolutions of $135\times240$ and $270\times480$, respectively. Let $\mathbf{f}^{\mathrm{dec}}_i$ denote the feature generated by the corresponding SNeRV and TAT blocks~\cite{zhang2024hnervboost}. The previous reconstruction $\mathbf{c}_t$ is resized and transformed into a scale-aligned context feature:

\begin{figure}[t]
	\centering
	\includegraphics[width=\columnwidth]{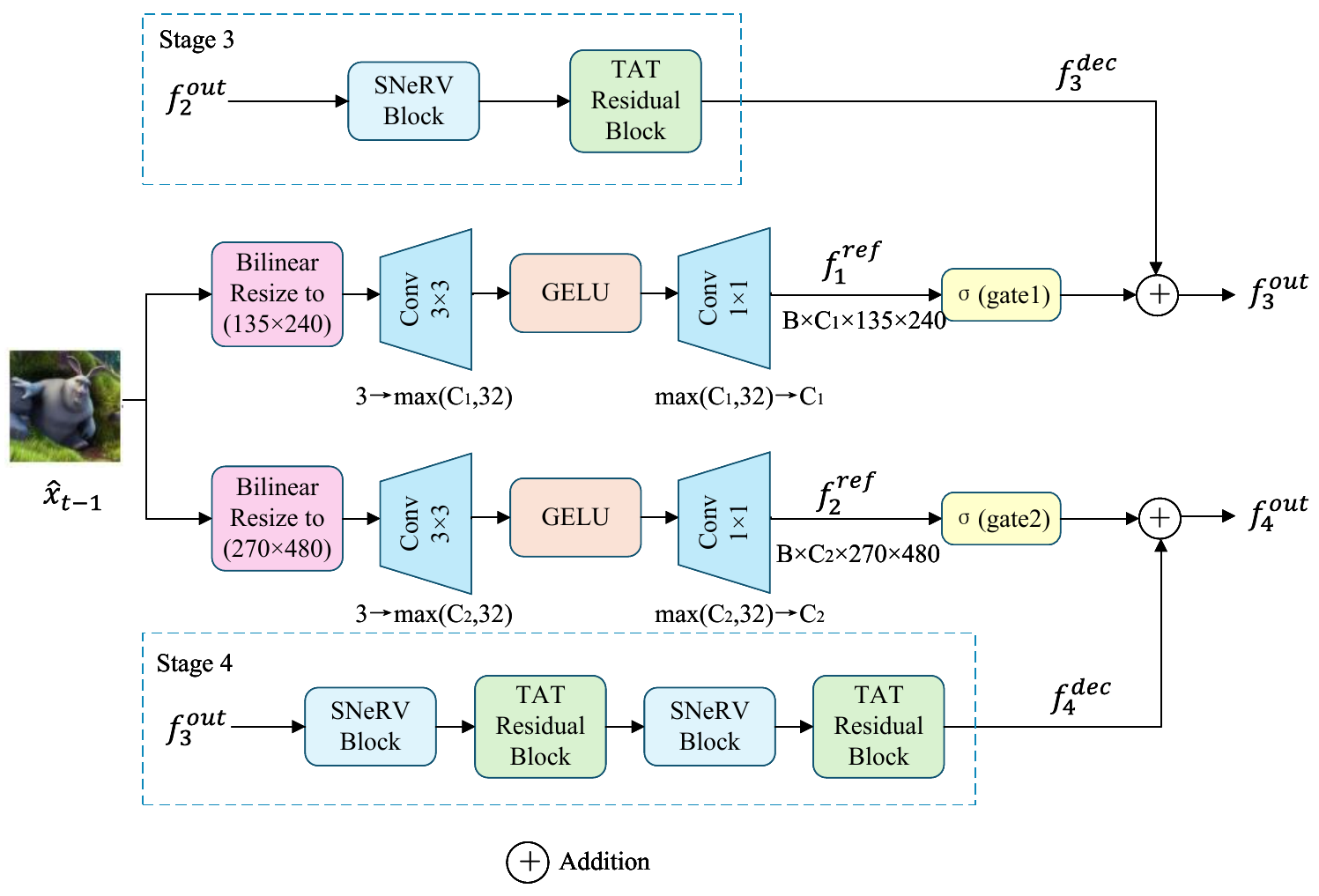}
	\caption{Architecture of MTCF. It leverages multi-level temporal context to enhance reconstruction and suppress unreliable historical information.}
	\label{fig:mtcf}
\end{figure}

\begin{table*}[t]
	\caption{Video representation results on the UVG dataset. Six methods are compared at three model scales in terms of parameters, computation, and PSNR.}
	\label{tab:uvg}
	\centering
	\renewcommand{\arraystretch}{0.9}
	\setlength{\tabcolsep}{1.8pt}
	\begin{tabular*}{\textwidth}{@{\extracolsep{\fill}}l*{10}{r}}
		\hline
		Model & Size & MACs & Beauty & Bosph. & Honey. & Jockey & Ready. & Shake. & Yacht. & Avg.\\
		\hline
		NeRV~\cite{chen2021nerv} & 3.31M & 227G & 32.83 & 32.20 & 38.15 & 30.30 & 23.62 & 33.24 & 26.43 & 30.97\\
		E-NeRV~\cite{li2022enerv} & 3.29M & 230G & 33.13 & 33.38 & 38.87 & 30.61 & 24.53 & 34.26 & 26.87 & 31.66\\
		HNeRV~\cite{chen2023hnerv} & 3.26M & 175G & 33.56 & 35.03 & 39.28 & 31.58 & 25.45 & 34.89 & 28.98 & 32.68\\
		HiNeRV~\cite{kwan2023hinerv} & 3.19M & 181G & \textbf{34.08} & 38.58 & \textbf{39.71} & 36.10 & 31.53 & 35.85 & 30.95 & 35.26\\
		HNeRV-Boost~\cite{zhang2024hnervboost} & 3.05M & 131G & 33.80 & 36.12 & 39.64 & 34.29 & 28.13 & 35.88 & 29.32 & 33.88\\
		TCNeRV & 3.06M & 131G & 34.06 & \textbf{39.68} & 39.62 & \textbf{36.49} & \textbf{33.58} & \textbf{36.02} & \textbf{33.11} & \textbf{36.08}\\
		\hline
		NeRV & 6.53M & 228G & 33.67 & 34.83 & 39.00 & 33.34 & 26.03 & 34.39 & 28.23 & 32.78\\
		E-NeRV & 6.54M & 245G & 33.97 & 35.83 & 39.75 & 33.56 & 26.94 & 35.57 & 28.79 & 33.49\\
		HNeRV & 6.40M & 349G & 33.99 & 36.45 & 39.56 & 33.56 & 27.38 & 35.93 & 30.48 & 33.91\\
		HiNeRV & 6.49M & 368G & \textbf{34.33} & 40.37 & \textbf{39.81} & \textbf{37.93} & 34.54 & \textbf{37.04} & 32.94 & 36.71\\
		HNeRV-Boost & 5.01M & 288G & 34.14 & 37.87 & 39.74 & 35.84 & 30.36 & 36.71 & 30.77 & 35.06\\
		TCNeRV & 5.04M & 289G & 34.21 & \textbf{40.38} & 39.74 & 37.34 & \textbf{34.89} & 36.74 & \textbf{34.03} & \textbf{36.76}\\
		\hline
		NeRV & 13.01M & 230G & 34.15 & 36.96 & 39.55 & 35.80 & 28.68 & 35.90 & 30.39 & 34.49\\
		E-NeRV & 13.02M & 285G & 34.25 & 37.61 & 39.74 & 35.45 & 29.17 & 36.97 & 30.76 & 34.85\\
		HNeRV & 12.87M & 701G & 34.30 & 37.96 & 39.73 & 35.47 & 29.67 & 37.16 & 32.31 & 35.23\\
		HiNeRV & 12.82M & 718G & \textbf{34.66} & \textbf{41.83} & \textbf{39.95} & \textbf{39.01} & \textbf{37.32} & \textbf{38.19} & 35.20 & \textbf{38.02}\\
		HNeRV-Boost & 10.03M & 700G & 34.42 & 39.75 & 39.83 & 37.57 & 33.12 & 37.85 & 32.90 & 36.49\\
		TCNeRV & 10.08M & 702G & 34.35 & 41.31 & 39.82 & 38.17 & 36.42 & 37.76 & \textbf{35.37} & 37.60\\
		\hline
	\end{tabular*}
\end{table*}

\begin{equation}
	\mathbf{f}^{\mathrm{ctx}}_i=
	\Phi_i\!\left[\operatorname{Resize}\!\left(\mathbf{c}_t;H_i,W_i\right)\right],
	\label{eq:mtcf-ctx}
\end{equation}
where $(H_3,W_3)=(135,240)$ and $(H_4,W_4)=(270,480)$. $\Phi_i$ comprises a $3\times3$ convolution, a GELU activation, and a $1\times1$ convolution, mapping the three input channels through $\max(C_i,32)$ channels to $C_i$ channels. Learnable gates then control context fusion:
\begin{equation}
	\mathbf{a}_i=\operatorname{sigmoid}(\mathbf{g}_i),\qquad
	\mathbf{f}^{\mathrm{out}}_i=
	\mathbf{f}^{\mathrm{dec}}_i+
	\mathbf{a}_i\odot\mathbf{f}^{\mathrm{ctx}}_i,
	\label{eq:mtcf-fusion}
\end{equation}
where $i\in\{3,4\}$ and $\odot$ denotes channel-wise multiplication. Stage 3 supplies layout and motion cues; Stage 4 enhances edges and textures; and the gates suppress unreliable context without altering the backbone structure.

\begin{figure}[t]
	\centering
	\includegraphics[width=\columnwidth]{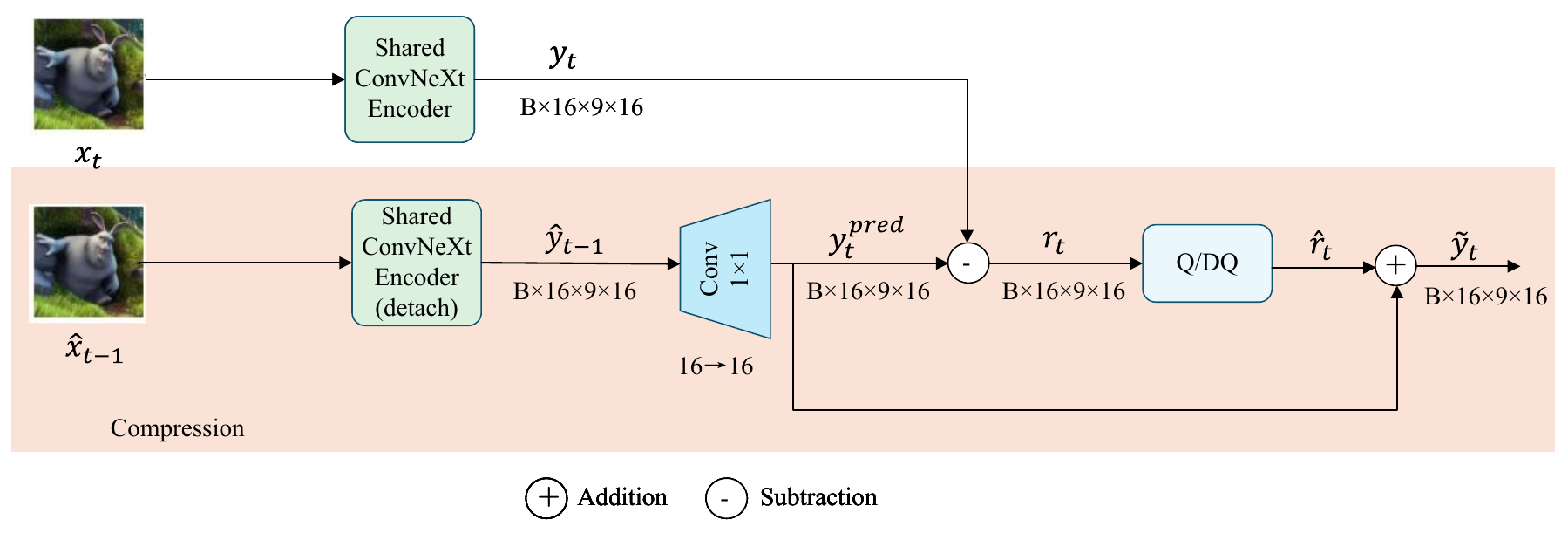}
	\caption{Architecture of TERC. It predicts the embedding from historical context, codes the prediction residual, and thus establishes explicit frame-level predictive coding prior to quantization.}
	\label{fig:terc}
\end{figure}

\subsection{Temporal Embedding-Residual Coding (TERC)}
\label{sec:terc}

Frame-wise embeddings determine reconstructed content and contribute to the bitstream. Existing INR codecs encode them independently, leaving temporal redundancy unexploited. TERC instead predicts each embedding from the historical reconstruction and encodes only its residual.

\begin{figure*}[t]
	\centering
\includegraphics[width=0.95\textwidth]{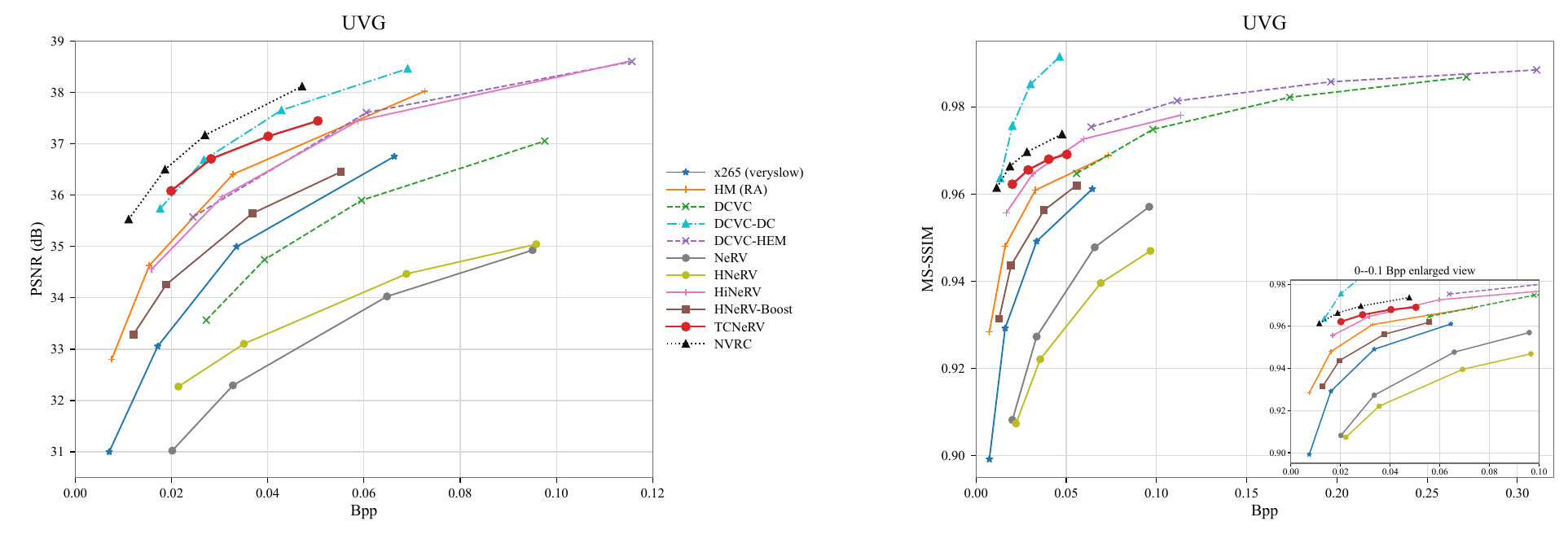}
	\caption{Rate-distortion comparison on the UVG dataset in terms of Bpp--PSNR and Bpp--MS-SSIM.}
	\label{fig:rd}
\end{figure*}

As shown in Fig.~\ref{fig:terc}, the shared ConvNeXt encoder~\cite{liu2022convnext} extracts the target embedding $\mathbf{y}_t=\mathcal{E}(\mathbf{x}_t;\phi)$, where $C_e=16$ and $(H_e,W_e)=(9,16)$. The current embedding is predicted as
\begin{equation}
	\mathbf{y}^{\mathrm{pred}}_t=
	\begin{cases}
		\mathbf{0}, & t=0,\\
		\mathcal{P}\!\left(\operatorname{stopgrad}\!\left[\mathcal{E}(\mathbf{c}_t;\phi)\right]\right), & t\geq1.
	\end{cases}
	\label{eq:terc-pred}
\end{equation}
where $\mathcal{P}$ is a $1\times1$ convolutional predictor. TERC quantizes and entropy codes only the residual $\mathbf{r}_t=\mathbf{y}_t-\mathbf{y}^{\mathrm{pred}}_t$. The decoder reconstructs the embedding as
\begin{equation}
	\widetilde{\mathbf{y}}_t=
	\operatorname{stopgrad}\!\left(\mathbf{y}^{\mathrm{pred}}_t\right)
	+\widehat{\mathbf{r}}_t.
	\label{eq:terc-recon}
\end{equation}
The first frame therefore codes the complete embedding, while subsequent frames encode only temporally unpredictable information. Stop-gradient prevents cancellation between the prediction and residual branches. The training loss, entropy regularization, and compression fine-tuning follow HNeRV-Boost~\cite{zhang2024hnervboost}.

\section{Experiments}
\label{sec:experiments}

\subsection{Experimental Configuration}
\label{sec:configuration}

We use the UVG~\cite{mercat2020uvg} dataset, comprising seven $1920\times1080$ videos with 600 or 300 frames. Regression performance is evaluated using PSNR and MS-SSIM. Compression performance is characterized by Bpp together with PSNR and MS-SSIM, and BD-rate~\cite{bjontegaard2001calculation} is computed over the common quality range. We adopt HNeRV-Boost~\cite{zhang2024hnervboost} with a ConvNeXt encoder~\cite{liu2022convnext} as the baseline. The base and embedding channel dimensions are 64 and 16, respectively. The upsampling scales are $[5,3,2,2,2]$, and the last five stages contain $[1,1,2,2,2]$ blocks. Regression models are trained for 300 epochs with Adan~\cite{xie2024adan}, a batch size of 1, an initial learning rate of $3\times10^{-3}$, 10\% warm-up, and cosine decay. Compression models are fine-tuned for 100 epochs at $5\times10^{-4}$ using 8-bit quantization, a global Gaussian entropy model, a rate-distortion coefficient of 0.05, and a target bit depth of 4 bits. Comparisons include x265~\cite{x265}, HM-18.0~\cite{rosewarne2022hm}, DCVC variants~\cite{li2021dcvc,li2023diverse,li2022hem}, and INR codecs~\cite{chen2021nerv,chen2023hnerv,kwan2023hinerv,zhang2024hnervboost,kwan2024nvrc}. Our neural models are implemented in PyTorch and trained on NVIDIA TITAN RTX GPUs.

\begin{table}[t]
	\caption{BD-rate (\%) of TCNeRV relative to the comparison methods on the UVG dataset.}
	\label{tab:bdrate}
	\centering
	\renewcommand{\arraystretch}{0.9}
	\setlength{\tabcolsep}{3pt}
	\begin{tabular*}{\columnwidth}{@{\extracolsep{\fill}}lrr}
		\hline
		Method & PSNR & MS-SSIM\\
		\hline
		x265 (veryslow)~\cite{x265} & $-58.76\%$ & N/A\\
		HM (RA)~\cite{rosewarne2022hm} & $-22.06\%$ & $-42.83\%$\\
		DCVC~\cite{li2021dcvc} & $-66.73\%$ & $-44.51\%$\\
		DCVC-DC~\cite{li2023diverse} & $8.31\%$ & $114.27\%$\\
		DCVC-HEM~\cite{li2022hem} & $-24.10\%$ & N/A\\
		HNeRV-Boost~\cite{zhang2024hnervboost} & $-57.26\%$ & N/A\\
		HiNeRV~\cite{kwan2023hinerv} & $-29.85\%$ & $-8.74\%$\\
		NVRC~\cite{kwan2024nvrc} & $40.22\%$ & $68.61\%$\\
		\hline
	\end{tabular*}
\end{table}

\subsection{Video Regression}
\label{sec:regression}

Table~\ref{tab:uvg} reports the results on the UVG~\cite{mercat2020uvg} dataset. With approximately 3M parameters, TCNeRV achieves 36.08 dB, exceeding HNeRV-Boost and HiNeRV by 2.20 and 0.82 dB, respectively. The corresponding gains over HNeRV-Boost at the medium and large scales are 1.70 and 1.11 dB, respectively. Larger improvements on high-motion sequences confirm the benefit of historical context.

\subsection{Video Compression}
\label{sec:compression}

Fig.~\ref{fig:rd} and Table~\ref{tab:bdrate} present the compression results on the UVG~\cite{mercat2020uvg} dataset. TCNeRV reduces BD-rate by 22.06\%, 66.73\%, 24.10\%, 57.26\%, and 29.85\% relative to HM, DCVC, DCVC-HEM, HNeRV-Boost, and HiNeRV, respectively. Its BD-rates relative to DCVC-DC and NVRC are $+8.31\%$ and $+40.22\%$, indicating remaining performance gaps.

\begin{table}[t]
	\caption{Ablation study of TCNeRV on the UVG dataset.}
	\label{tab:ablation}
	\centering
	\renewcommand{\arraystretch}{0.9}
	\setlength{\tabcolsep}{2.5pt}
	\begin{tabular*}{\columnwidth}{@{\extracolsep{\fill}}lccrrr}
		\hline
		Model & MTCF & TERC & Size & MACs & Avg.\\
		\hline
		Baseline~\cite{zhang2024hnervboost} & & & 3.05M & 130.6G & 33.88\\
		& $\checkmark$ & & 3.06M & 131.0G & 35.96\\
		& & $\checkmark$ & 3.05M & 130.6G & 34.08\\
		Ours& $\checkmark$ & $\checkmark$ & 3.06M & 131.0G & 36.08\\
		\hline
	\end{tabular*}
\end{table}

\subsection{Ablation Study}
\label{sec:ablation}

Table~\ref{tab:ablation} shows that MTCF improves the average PSNR from 33.88 to 35.96 dB while adding only 0.01M parameters and 0.4G MACs. TERC alone provides a 0.20 dB improvement, and the full model reaches 36.08 dB, exceeding the MTCF-only variant by 0.12 dB. These results indicate that MTCF provides the primary reconstruction gain, while TERC offers a modest additional improvement.

\section{Conclusion}
\label{sec:conclusion}

TCNeRV uses historical reconstructions for gated multi-scale decoding-feature fusion and content-embedding residual coding. With approximately 3M parameters, it reduces BD-rate by $57.26\%$ relative to HNeRV-Boost, demonstrating competitiveness among implicit neural video codecs. Future work will address long-sequence errors, scene changes, random access, and rate-model optimization.

\begingroup
\renewcommand{\baselinestretch}{0.85}\selectfont
\let\standardthebibliography\thebibliography
\renewcommand{\thebibliography}[1]{%
	\standardthebibliography{#1}%
	\setlength{\topsep}{0pt}%
	\setlength{\partopsep}{0pt}%
	\setlength{\itemsep}{0pt}%
	\setlength{\parsep}{0pt}%
}
\bibliographystyle{IEEEbib}
\bibliography{paper_icassp2027_full}
\endgroup

\end{document}